\documentclass[letterpaper, 10 pt, conference]{ieeeconf}  

\IEEEoverridecommandlockouts                              

\usepackage{amssymb}  
\usepackage{mathtools}
\usepackage{cite}
\usepackage{bm}
\usepackage{textcomp}
\usepackage{siunitx}
\usepackage{comment}
\usepackage{amsmath}
\usepackage{amsfonts}
\usepackage{subfig}
\usepackage{caption}
\usepackage{float}

\DeclareMathOperator*{\argmax}{argmax}

\DeclarePairedDelimiterX{\infdivx}[2]{(}{)}{%
  #1\;\delimsize|\delimsize|\;#2%
}
\usepackage{algorithm}
\usepackage{algorithmic}

\def\BibTeX{{\rm B\kern-.05em{\sc i\kern-.025em b}\kern-.08em
    T\kern-.1667em\lower.7ex\hbox{E}\kern-.125emX}}
\usepackage{balance}
\usepackage{booktabs}
\usepackage{graphicx}
\usepackage[normalem]{ulem}
\usepackage{etoolbox}
\usepackage{hyperref}
\usepackage{color}
\usepackage{todonotes}

\title{\LARGE \bf
Reinforcement Learning with Evolutionary Trajectory Generator: A General Approach for Quadrupedal Locomotion
}

\vspace{-.2cm}
\author{Haojie Shi$^{1*}$, Bo Zhou$^{2*}$, Hongsheng Zeng$^2$, Fan Wang$^{2\dag}$, \\ Yueqiang Dong$^2$, Jiangyong Li$^2$, Kang Wang$^2$, Hao Tian$^2$, Max Q.-H. Meng$^{3\dag}$, \textit{Fellow, IEEE} }

\begin{document}
\twocolumn[{%
\renewcommand\twocolumn[1][]{#1}%

\maketitle
\thispagestyle{empty}
\pagestyle{empty}
\vspace{-.3cm}
\begin{figure}[H]
\hsize=\textwidth
    \centering
    \includegraphics[width=\textwidth]{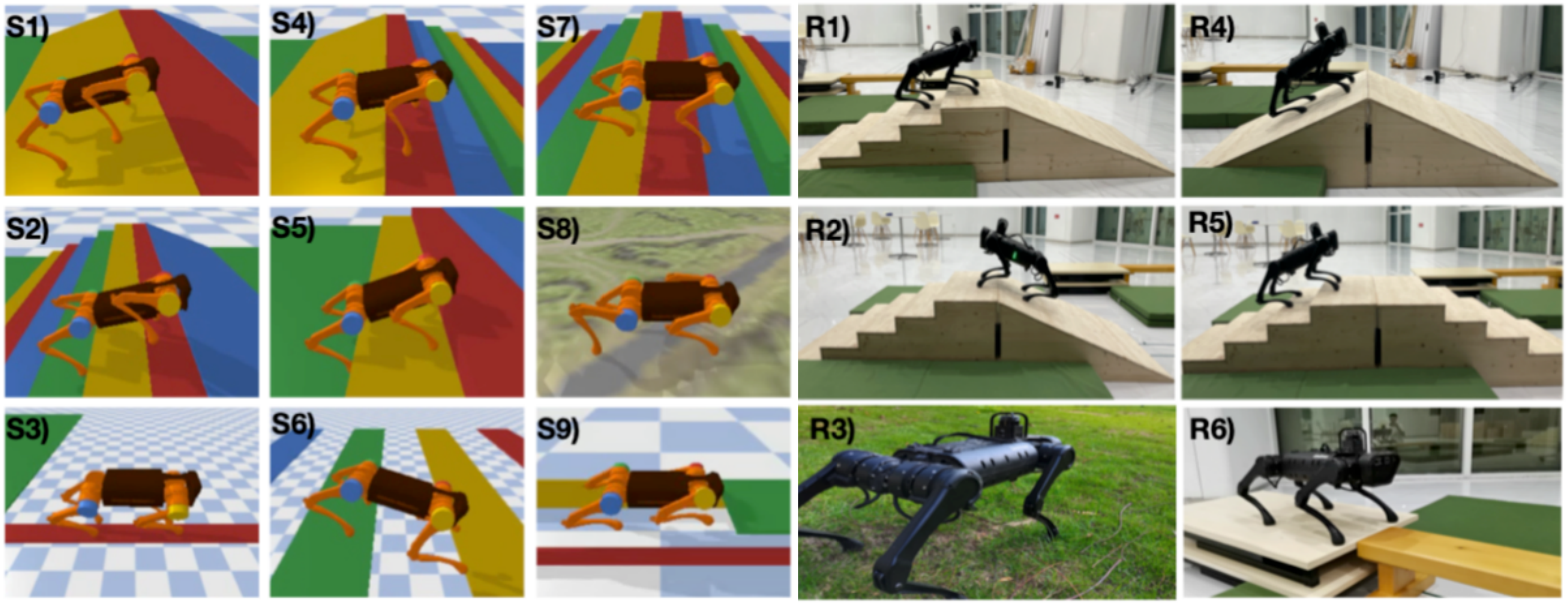}
    \caption{The proposed approach can adapt the quadrupedal robot to a wide range of scenarios, including 9 simulation tasks and 6 real-world tasks. (S1) SlopeSlope: Climbing up and down slopes. (S2) StairSlope. (S3) Balance: Walking over balance beam(width=12cm). (S4) SlopeStair. (S5) Stair13: Climbing up higher stair (width=40cm, height=13cm). (S6) Gallop: Gallop over planks with wide gap (width=50cm). (S7) StairStair. (S8) Terrain. (S9) Cave: Crawling through narrow cave (height=17cm). (R1) StairStair. (R2)  SlopeStair. (R3) Terrain: Walking across uneven terrain. (R4) SlopeSlope. (R5) StairSlope. (R6) Balance: Walking over balance beam (width=20cm). Note that except Stair13, the stair height for all simulation tasks is 8 cm, and the stair width is 25 cm. The slope angle in all tasks is 20$^{\circ}$. }
    \label{fig:env}
\end{figure}
\vspace{-.3cm}
}]
\begin{abstract}
\let\thefootnote\relax\footnotetext{*  Equal Contribution.}
\let\thefootnote\relax\footnotetext{$\dag$ Corresponding author.}
\let\thefootnote\relax\footnotetext{1 Haojie Shi is from the Chinese University of Hong Kong, Hong Kong, (email: h.shi@link.cuhk.edu.hk)}
\let\thefootnote\relax\footnotetext{2 Bo Zhou, Hongsheng Zeng, Fan Wang, Yueqiang Dong, Jiangyong Li, Kang Wang and Hao Tian are affiliated with Baidu Inc., China. (email: {zhoubo01,  zenghongsheng, wang.fan, dongyueqiang, lijiangyong01, wangkang02, tianhao}@baidu.com)}
\let\thefootnote\relax\footnotetext{3 Max Q.-H. Meng is with the Department of Electronic and Electrical
Engineering of the Southern University of Science and Technology in
Shenzhen, China, on leave from the Department of Electronic Engineering,
the Chinese University of Hong Kong, Hong Kong, and also with the
Shenzhen Research Institute of the Chinese University of Hong Kong in
Shenzhen, China. (email: max.meng@cuhk.edu.hk)}
Recently reinforcement learning (RL) has emerged as a promising approach for quadrupedal locomotion, which can save the manual effort in conventional approaches such as designing skill-specific controllers. However, due to the complex nonlinear dynamics in quadrupedal robots and reward sparsity, it is still difficult for RL to learn effective gaits from scratch, especially in challenging tasks such as walking over the balance beam. To alleviate such difficulty, we propose a novel RL-based approach that contains an evolutionary foot trajectory generator. Unlike prior methods that use a fixed trajectory generator, the generator continually optimizes the shape of the output trajectory for the given task, providing diversified motion priors to guide the policy learning. The policy is trained with reinforcement learning to output residual control signals that fit different gaits. We then optimize the trajectory generator and policy network alternatively to stabilize the training and share the exploratory data to improve sample efficiency. As a result, our approach can solve a range of challenging tasks in simulation by learning from scratch, including walking on a balance beam and crawling through the cave. To further verify the effectiveness of our approach, we deploy the controller learned in the simulation on a 12-DoF quadrupedal robot, and it can successfully traverse challenging scenarios with efficient gaits.\footnotemark[1]

\footnotetext[1]{We provide a video to show the learned gaits in different tasks in YouTube: \href{https://www.youtube.com/watch?v=hgBLR09MEOw}{youtube.com/watch?v=hgBLR09MEOw}, and code is available in Github: \href{https://github.com/PaddlePaddle/PaddleRobotics/tree/main/QuadrupedalRobots/ETGRL}{github.com/PaddlePaddle/PaddleRobotics} }
\end{abstract}

\section{INTRODUCTION}

The research and development of the quadrupedal robot have attracted much attention due to its potential to explore various terrains. Several approaches have achieved versatile locomotion skills in the quadrupedal robot, by designing the foot trajectories and the corresponding controller manually \cite{hyun2014high,gonzalez2020line,iqbal2020provably}. However, the performance of the designed controller relies on the expert knowledge of both the robotics system and the desired locomotion skill, and the development pipeline often involves tedious manual tuning \cite{tan2018sim,peng2020learning}.

Recently reinforcement learning has emerged as a promising approach to save the manual effort in conventional approaches, learning the controller from scratch without any domain knowledge \cite{kohl2004policy,lee2020learning,yang2020multi}. While RL-based controllers have achieved promising results in a variety of tasks, learning efficient quadrupedal locomotion skills from scratch is still challenging, due to the highly non-linearity in dynamics.
To address the problem, previous work employs a trajectory generator to provide motion prior to guide the policy learning \cite{lee2020learning,tan2018sim}. Despite the progress in learning-based control of locomotion skills, an unified framework for learning versatile skills remains an open problem for quadrupedal robots, especially in challenging environments (e.g., walking on a narrow balance beam).

In this work, we propose a novel RL-based approach that can adapt to diverse challenging tasks for quadrupedal robots. Following prior work \cite{lee2020learning,tan2018sim}, our controller consists of two components: an open-loop foot trajectory generator (FTG) generating time-dependent foot trajectory, and a neural network (NN) based policy generating residual control signals. The trajectory generator allows us to provide priors on motion generation. However, previous works adopt fixed FTGs, providing only pre-fixed basic trajectories that may not fit specific tasks. For challenging tasks, it still requires much manual effort to provide reasonable motion priors. In case the provided motion prior is inefficient for the given task, it might even deteriorate the performance of RL-based policy. To overcome the limitation, we propose to simultaneously optimize the FTG and the policy through evolutionary strategies (ES) \cite{salimans2017evolution} and RL algorithms, respectively. We employ ES to search the optimal foot trajectory in the trajectory space, efficiently optimizing the motion prior, instead of searching in parameter space like prior works.
The evolved FTG can thus generate patterns better adapted to the specific task, resulting in more effective learning of RL-based policies. 
Finally, since the two components have to couple with each other to accomplish various tasks, to stabilize the training, we further propose to alternatively optimize the trajectory generator and the NN-based policy, freezing the parameters of one component while training the other. 

We evaluate our approach in a range of challenging simulation tasks, such as walking on a balance beam and jumping across plank intervals (Figure~\ref{fig:env}). Results prove the proficiency of our approach compared with the other previous learning-based proposals, while most of them cannot make much progress in those challenging tasks.
We further deploy the controller on a 12-DoF quadrupedal robot by applying sim-to-real transferring, demonstrating the potential of applying the proposed framework in real-world tasks.

The main contributions of this paper are: 
\begin{enumerate}
  \item We devise a novel paradigm to optimize the foot trajectories generated by the FTG, providing motion priors to guide the policy learning.
  \item We propose a novel learning-based approach for quadrupedal locomotion that alternately optimizes trajectory generator via ES and a neural network policy via RL, capable of learning efficient gaits in challenging tasks.
  \item We successfully transfer the controller learned in the simulation to a 12-DoF real quadrupedal robot and demonstrate its effectiveness in challenging tasks, such as climbing up stairs and walking over a balance beam.
\end{enumerate}

\section{RELATED WORK}
A lot of effort has been put into designing the controller for the quadrupedal robot since it holds the promise of traversing complex environments such as stairs and irregular terrains. The most intuitive direction is to mimic animal behaviors and reproduce their locomotion skills in quadrupedal robots. This motivates the development of the trajectory generator that provides the predefined foot trajectory to control quadrupedal robots. While manually designed trajectory generators have successfully reproduced the animal behaviors in robotics, they lack the ability to adapt the controller to the environment, and learning-based approaches attract much interest due to their ability to adjust the controller behavior during the interaction with the environment. In this section, we briefly review the trajectory generator and learning-based approaches for quadrupedal locomotion.


\textbf{Open-loop Trajectory Generator.} 
Sine curves are often used to generate cycle signals. In \cite{tan2018sim}, the desired motors angles of each leg are decided by two sine curves. One of the curves outputs the signal for swing while the other is for the extension signal. \cite{paigwar2020robust} explores walking arbitrary slopes with the semi-elliptic foot trajectory generator by estimating the terrain slope in real-time and dynamically shaping the trajectory. Central pattern generator (CPG) \cite{ijspeert2008central} was first found in animals that can produce rhythmic patterns of neural activity without any rhythmic input signal. Motivated by this nature mechanism, researchers have designed a a variable trajectory generators that can mimic the locomotion skills of animals, including fish robots \cite{crespi2008controlling,wang2014cpg}, snake-like robots \cite{kamimura2005automatic,wu2017cpg} and legged robots \cite{endo2008learning,liu2011cpg,pinto2012stability}.

\textbf{Learning-based approach.} In recent years the success of learning-based strategies provides an alternative approach for learning efficient quadrupedal locomotion skills. \cite{lee2020learning} learns a policy network via RL to control the quadrupedal robot to traverse challenging terrains, with a neural policy network that performs the action on a stream of proprioceptive signals. \cite{kumar2021rma} introduces an adaptation module to facilitate fast adaption on uneven terrains, by predicting the environmental factors based on the history of states and actions. \cite{yang2020multi} proposes a multi-expert learning architecture that contains a set of expert neural networks and a gating neural network for knowledge fusion. The expert networks are first trained separately to learn unique motor skills, and the gating network is then trained to fuse the expert networks to generate adaptive skills. In \cite{tan2018sim}, the controller consists of an open-loop controller and a trainable neural network that decides the output based on the observation. The control signal is the sum of two output signals given by the two components. Our controller adopts the same architecture as \cite{tan2018sim}, in which the trajectory generator provides reference trajectories for policy learning. However, we improve the trajectory generator to an adaptive generator that optimizes its output trajectories according to the given task, providing efficient motion priors to guide the policy learning.

\section{PRELIMINARIES}
In this section, we first formulate the task of controlling quadrupedal robots as a Markov decision process (MDP) and introduce solving the problem via RL. Since our controller contains a trajectory generator to provide motion prior for policy learning, we then introduce CPG for quadrupedal locomotion.
\subsection{Deep reinforcement learning for quadrupedal locomotion}

We now formulate the task of designing a learning-based controller for quadrupedal locomotion as a MDP defined by the tuple $( \mathcal{S}, \mathcal{A}, P, r, p )$, where $\mathcal{S}$ is a set of all possible states, $\mathcal{A}$ is a set of actions, $P(s_{t+1}|s_t, a_t)$ is a probabilistic transition function, $r(s_t, a_t)$ is a reward function, and $p(s_0)$ represents the distribution of the initial state. At each time step t, the controller provides control signal $a_t \in \mathcal{A}$ for the robot according to the state $s_t \in \mathcal{S}$. 
As the robot moves with the control signal $a_t$, it can receive a reward $r_t$, and the next state $s_{t+1}$ depends on the state transition function $P(s_{t+1}|s_t, a_t)$. Denoted by $\pi(s_t)$, the goal of the learning-based controller is to find an optimal controller $\pi^*$ that can maximize the average reward it receives: 
\begin{equation}
    \pi^*=\argmax_{\pi} \mathbb{E}_{a_t\sim \pi,s_{t+1}\sim P} \sum_{t=0}^\infty \gamma^t r_t, 
    \label{eq:rl_object}
\end{equation}
where $\gamma \in (0,1)$ is the discount factor for accumulative reward computation.

In this work, we adopt the actor-critic reinforcement learning approach \cite{haarnoja2018soft} to find the optimal controller. We first define a state-action value function $Q_{\phi}(s_t,a_t)$ parameterized by a neural network $\phi$ to estimate the expected future reward:
\begin{equation}
Q_{\phi}(s_t,a_t) = \mathbb{E}_{a_t \sim \pi(s_t)} \sum_{t=0}^{\infty} \gamma^t r_t.
\label{eq:Q_definition}
\end{equation}
The state-action value function can be approximated by minimizing the temporal difference (TD) error \cite{sutton2018reinforcement}:
\begin{align}
\label{eq:TD_loss}
\mathcal{L}(Q_{\phi}) = (Q_{\phi}(s_t, a_t) &- (r_t + \gamma*Q_{\phi}(s_{t+1}, a_{t+1}))^2, \\
a_{t+1} &\sim \pi(s_{t+1}) \nonumber
\end{align}
Using a neural network $\theta$ to represent the policy $\pi_{\theta}(s_t)$, we can optimize the policy towards maximizing the expected future reward:
\begin{align}
L(\pi_{\theta}) = -Q_{\phi}(s_t, a_t), a_t \sim \pi_{\theta}.
\label{eq:policy_optimization}
\end{align}
Here we represent the loss function with the negative Q value function. To stabilize the training, prior work \cite{lillicrap2015continuous} uses an experience replay buffer \cite{mnih2015human} to store the transitions $(s_t, a_t, r_t, s_{t+1})$ for training, and introduces two target networks \cite{mnih2015human} for TD error computation that maintain a slow copy of $Q_{\phi}(s_t, a_t)$ and $\pi_{\theta}(s_t)$, respectively.

\subsection{Problem setting}

\subsubsection{State and action space}
The state $s_t$ includes roll, pitch, yaw obtained by the IMU, joint angle and angular velocity of each joint in the leg, four binary values indicating whether the leg contacts the ground, and estimated body velocity of the robot. The control signals $a_t$ represents the desired angle of each joint, and the desired angle is tracked by the joint-level PD controller.

\subsubsection{Reward function}
The reward function is designed to learn robust and efficient locomotion skills for quadrupedal robots, while reducing the energy consumption: 
\begin{align}
    r_t = (1 - \lambda) (P_{t-1} - P_{t}) \cdot d_t - \lambda * \Delta_e(t)
\end{align}
where $P_t$ represents the robot position, $d_t$ is the desired direction, and $\Delta_e(t)$ represents the consumed energy. The weight $\lambda \sim (0,1)$ decides the importance of energy consumption in the reward function. The first term encourages the robot to move along the desired direction as fast as possible, while the second discourages unnecessary energy consumption.

\subsection{CPG for locomotion controller}
CPG serves as a central module in animal locomotion that produces rhythmic signals, and it has been applied for quadrupedal controllers. We can use two sine waves to represent the swing and the extension of each leg\cite{tan2018sim}:
\begin{align}
    c_0(t) &= A*sin(\omega t) \\
    c_1(t) &= A*sin(\omega t + B)
    \label{eq:CPG_1}
\end{align}
where $A, \omega, B$ represent amplitude, frequency, and the phase between two signals, respectively. 



To mitigate the manual effort of designing the CPG for the specific task, CPG-RBF \cite{thor2020generic} leverages a radial basis function (RBF) network \cite{broomhead1988radial} to shape the foot trajectory according to the task. The experimental result in CPG-RBF shows that by adjusting the parameters in the RBF network, it can express a variety of foot trajectories. CPG-RBF employs black-box optimization to find the optimal foot trajectory. The RBF network contains only a hidden layer, and its output depends on the CPG output $c_0(t), c_1(t)$. Suppose the number of neurons in the hidden layer is $H$, and each neuron is computed by:
\begin{align}
    V_i(t) &= exp\left\{-\frac{(c_0(t)-\mu_{i,0})^2+(c_1(t)-\mu_{i,1})^2}{\sigma_{RBF}^2}\right\}, \\
    &i = 0,...,H-1 \nonumber
\end{align}
where $\mu_{i,0}$ and $ \mu_{i,1}$ are two means of RBF neuron, $\sigma_{RBF}^2$ is the variance between neuron outputs. $\mu_{i,0}$ and $ \mu_{i,1}$ are computed by: 
\begin{equation}
    \mu_{i,j} = c_j(\frac{(i-1)*T}{H-1}), j = 0,1
\end{equation}
where $T=\frac{2\pi}{\omega}$ is the period of CPG signals. The hidden layer is followed by a linear layer that generates the target joint position:
\begin{equation}
P = W*V+b
\label{eq:CPG-RBF-Linear}
\end{equation}
where $W, b$ are the parameters of the linear layer.

\begin{figure*}[ht]
\centering
\includegraphics[width=\textwidth]{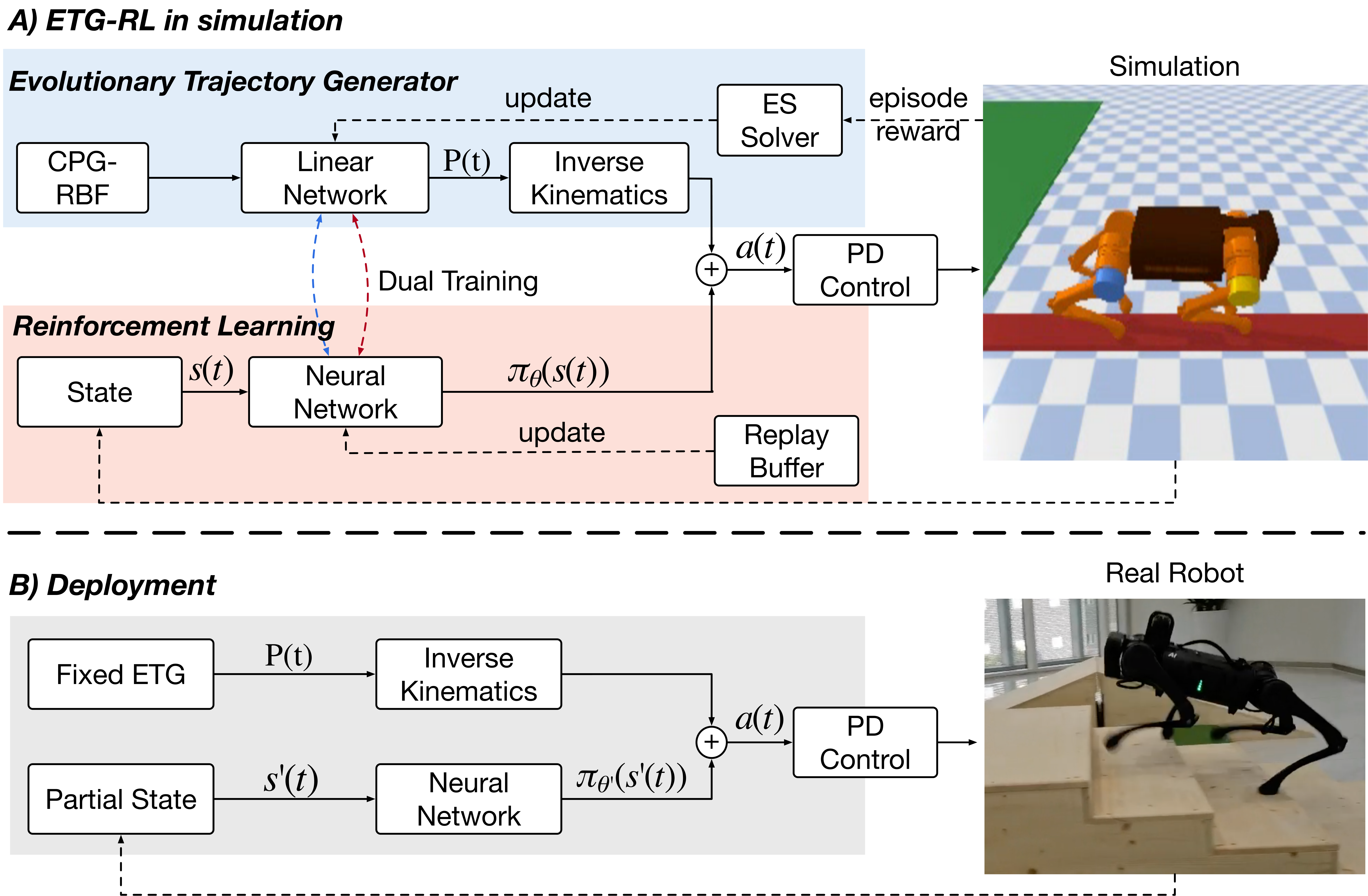}
\caption{The controller architecture of our approach. \textbf{A) Simulation training.} We alternative evolve the trajectory generator and optimize the policy. The trajectory generator can provide reference foot trajectories to guide policy optimization. \textbf{B) Real-world deployment.} We transfer the controller learned in simulation environments to the 12-DoF quadrupedal robot developed by Unitree Robotics.}
\label{fig:overview}
\vspace{-.3cm}
\end{figure*}
\section{Methodology}
We now introduce a novel RL-based approach for quadrupedal locomotion that can learn efficient gaits in challenging tasks. We adopt a similar architecture as the prior RL-based approaches \cite{lee2020learning,tan2018sim}, consisting of an open-loop trajectory generator and a neural network policy. The trajectory generator can provide periodic signals for control and motion priors to guide the policy learning \cite{tan2018sim,lee2020learning}.
We first improve the foot trajectory generator with an evolutionary trajectory generator (ETG), providing diversified reference motion trajectories for policy learning. We devise an optimization approach that directly searches at the trajectory space, which is more efficient than searching at the parameter space. It is easy to incorporate motion prior, by specifying the desired foot trajectory directly. The neural network policy is trained with reinforcement learning to acquire higher rewards in the simulation environment. We refer the proposed approach as ETG-RL, and the controller architecture is shown in Figure~\ref{fig:overview}.

\begin{figure}[ht]
\centering
\includegraphics[width=0.5\textwidth]{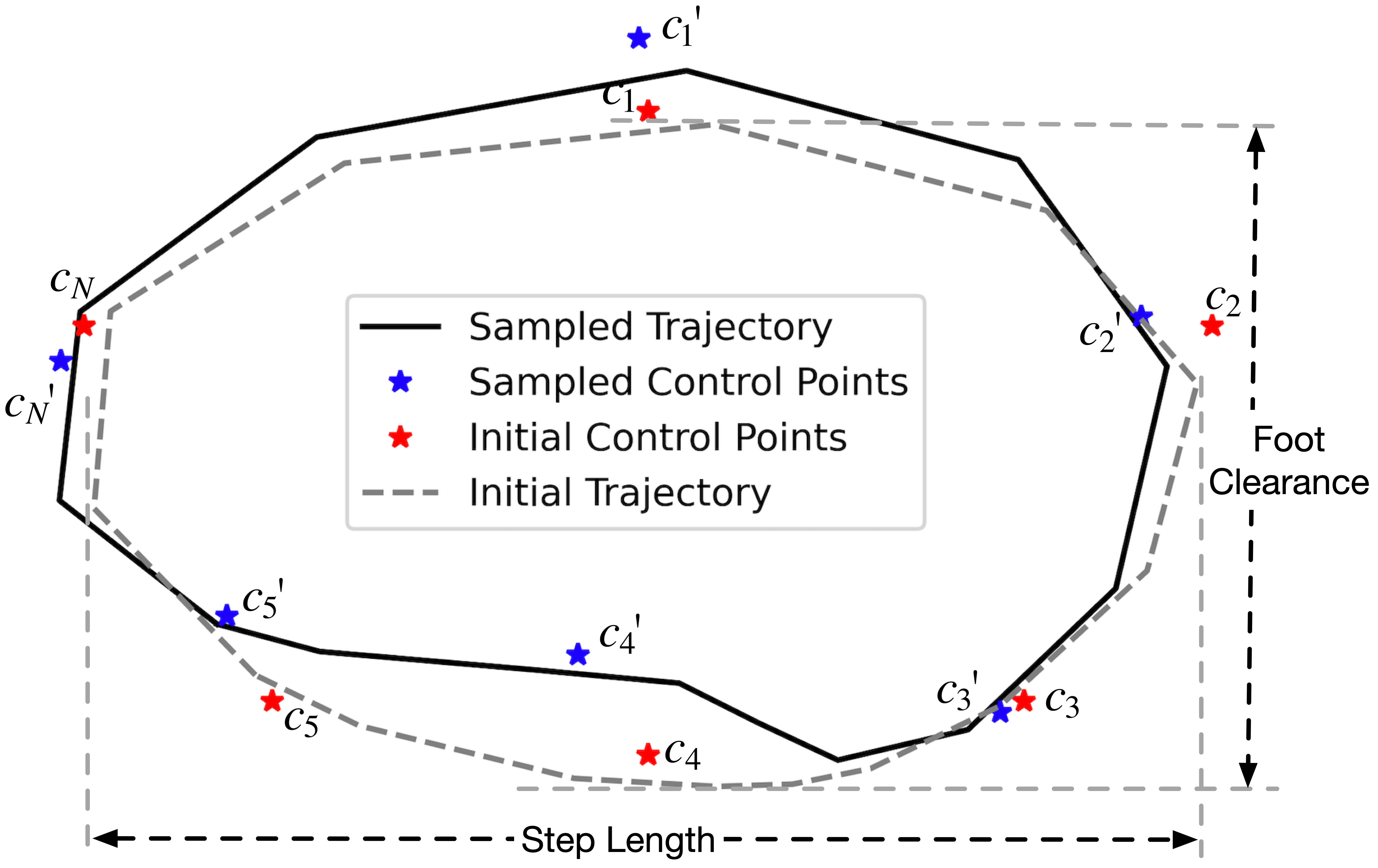}
\caption{ETG: Searching at the foot trajectory space. \textbf{Dotted line:} The desired trajectory represented by the red stars. \textbf{Solid line:} The sampled trajectory represented by the blue stars.}
\label{fig:ETG}
\end{figure}

\subsection{Evolutionary Trajectory Generator}
Trajectory generator allows us to provide motion prior to the controller, and the trajectories can also act as a reference for policy learning \cite{tan2018sim}. To provide diversified reference trajectories, we optimize the shape of the generated trajectory, following CPG-RBF \cite{thor2020generic}. While CPG-RBF introduces the RBF network to shape the trajectory, it is difficult to introduce the motion prior to the controller, because the trajectory is determined by the parameters in the RBF network, and it is difficult to specify the desired foot trajectory by setting the parameters directly. To address the problem, we directly perform the search at the trajectory space, which allows us to provide the motion prior for the optimization. 

As shown in Figure~\ref{fig:ETG}, the dotted line is the desired trajectory that we provide as the motion prior, and we sample $n$ control points (red stars) to represent the trajectory. Let $\tau_0$ denote the desired trajectories, and let $P(\tau_0)$ denote the set of corresponding $n$ control points. Now we have the target positions, and we can compute the parameters in the RBF network that generates the desired trajectory by solving Equation~(\ref{eq:CPG-RBF-Linear}) with time-dependent input $V$. The solution parameters $\phi_0 = (W_0, b_0)$ are used to initialize the RBF network.

We then search the optimal foot trajectory at the trajectory space with the evolutionary strategy \cite{salimans2017evolution} (ES). Searching at the trajectory space is more efficient than the parameter space, since it has a smaller dimension, and sampling in the parameter space may generate some unexpected trajectories that deviate from the current trajectory. At each iteration, we sample a number of exploration trajectories by adding Gaussian noises into the control points of the current trajectory (e.g., blue stars in Figure~\ref{fig:ETG} that forms the sampled trajectory). We then update the parameters of the RBF network according to the performance of each sampled trajectories, following \cite{salimans2017evolution}. The ETG algorithm is presented in Algorithm~\ref{alg:ETG}.

\begin{algorithm}[h]
    \caption{Evolutionary Trajectory Generator (ETG)}
    \label{alg:ETG}
    \begin{algorithmic}[1]
    \REQUIRE $\phi_0$, motion prior \\ n, number of control points \\ $\alpha$, learning rate
    \STATE initialize trajectory parameter: $\phi \leftarrow \phi_0$
    \WHILE{not converged}
      \STATE generate foot trajectory with $\phi$
      \STATE sample n control points from current foot trajectory
      \STATE add noise into the control points to generate K different trajectories
      \FOR{each k in \{1...K\}}
       \STATE compute corresponding $\phi_k$ with the sampled control points by solving Equation~\ref{eq:CPG-RBF-Linear}
       \STATE compute parameter distance: $\triangledown_{\phi_k} = \phi - \phi_k$
       \STATE evaluate the fitness of the sampled trajectory: $r_k$
      \ENDFOR
      \STATE gradient aggregation: $g=\sum_{k=1}^Kr_k\triangledown_{\phi_k}$
      \STATE update trajectory: $\phi \leftarrow \phi + \alpha * g$
    \ENDWHILE
    \end{algorithmic}
\end{algorithm}
\subsection{Policy learning via RL}
In addition to the open-loop trajectory generator, our controller contains a neural network policy $\pi_{\theta}(a_t)$ that outputs residual control signals. We adopt the SAC algorithm \cite{haarnoja2018soft} for policy optimization, an off-policy actor-critic algorithm that maximizes both the expected reward and entropy. The output of the policy is a vector representing the desired joint degrees of all the legs. As shown in Figure~\ref{fig:overview}, the policy output is combined with the output of the trajectory generator to form the final desired joint degree. We expand the input of the policy compared with prior approaches, and it contains both the sensor data and the control signal produced by the trajectory generator. The input expansion allows the policy to learn to output proper residual control signals for different foot trajectory generators, while the ETG algorithm continually optimizes the foot trajectory.
\subsection{Dual Training}
We optimize the trajectory generator and neural network policy separately and alternatively, which we call dual training. While optimizing the trajectory generator, we freeze the policy parameters and vice versa. The dual training can stabilize the training, since updating the trajectory generator or the policy results in a different behavior distribution. To further obtain diversified trajectories, we store the sampled trajectories in the optimization of ETG and append them into the experience replay buffer for reinforcement learning. The data expansion allows us to reuse the sampled foot trajectories of ETG for policy optimization, and thus improve sample efficiency. The complete algorithm is presented in Algorithm~\ref{alg:ETG-RL}.

\begin{algorithm}[h]
    \caption{ETG-RL Algorithm}
    \label{alg:ETG-RL}
    \begin{algorithmic}[1]
    \REQUIRE $\phi_0$, motion prior \\  $\mathcal{B}$, experience replay buffer \\ $\alpha$, learning rate \\ $E$, maximum train step for RL in each iteration \\ $\epsilon \sim \mathcal{N}(0,1)$, Gaussian noise for exploration
    \STATE Initialize the policy and trajectory generator.
    \WHILE{not converged}
    \STATE /*ETG phase*/
    \STATE optimize the trajectory shape with Algorithm~\ref{alg:ETG}
    \STATE append transitions $(s_t,a_t,r,s_{t+1})$ into $\mathcal{B}$
    \STATE /*RL phase*/
    \WHILE{not reach maximum train step $E$}
    \STATE rhythmic signal from ETG: $a_{ETG}(t)$
    \STATE residual control signal: $a_{RL}(t) = \pi_{\theta}(s) + \epsilon$
    \STATE generate control signal: $a(t) = a_{ETG}(t)+ a_{RL}(t)$ 
    \STATE append the transition $(s_t,a_t,r,s_{t+1})$ into $\mathcal{B}$
    \STATE update the Q value function with Equation~(\ref{eq:TD_loss})
    \STATE update the policy with Equation~(\ref{eq:policy_optimization})
    \ENDWHILE
    \ENDWHILE
    \end{algorithmic}
\end{algorithm}

\begin{figure*}[h]
\hsize=\textwidth
\centering
\includegraphics[width=\textwidth]{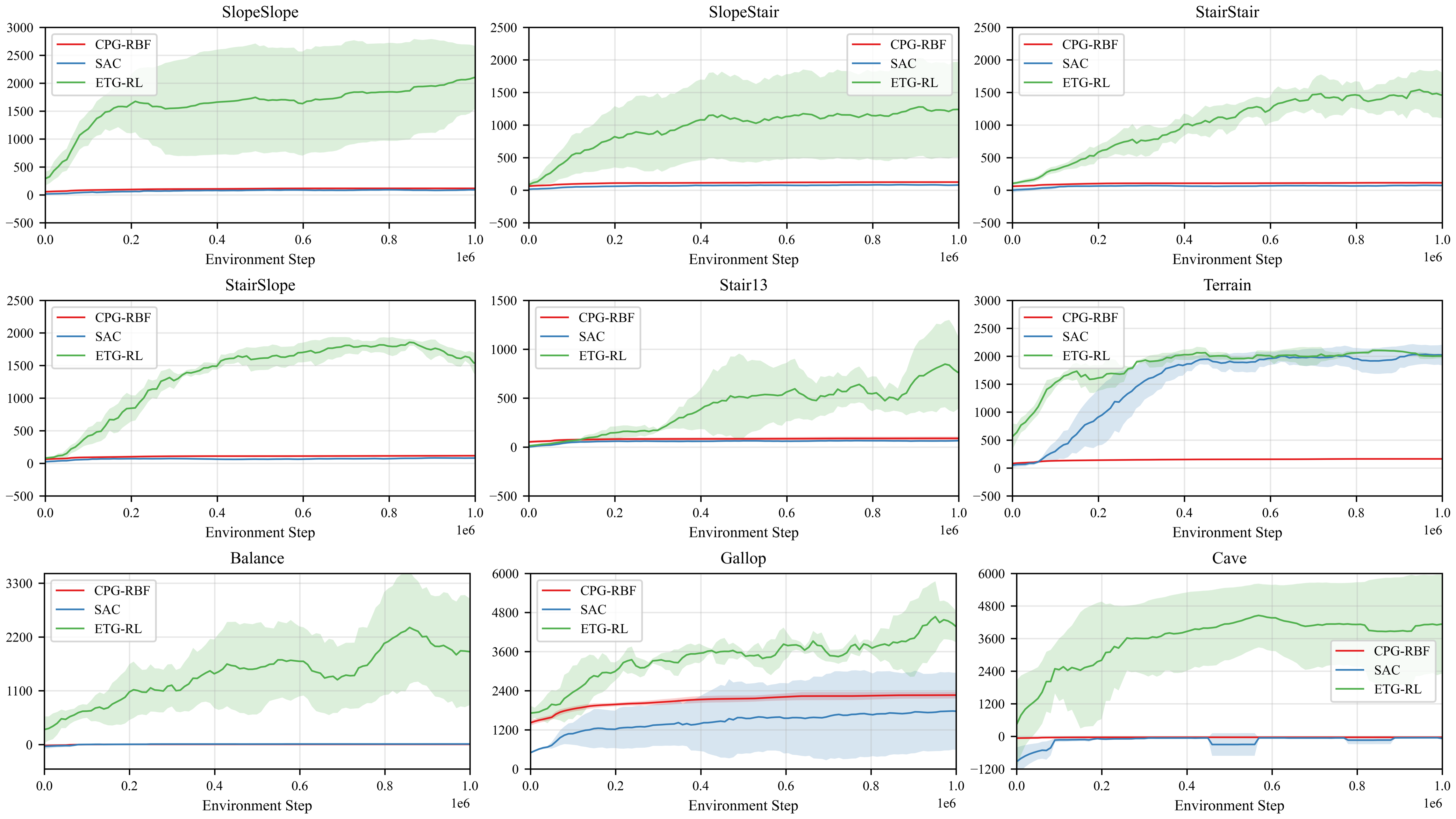}
\caption{Training curves of ETG-RL and the baseline methods on 9 simulation tasks. The 9 tasks correspond to scenarios S1-S9 in Figure~\ref{fig:env}, respectively. Each experiment was run four times with different random seeds. The solid lines represent the average score, while the shaded areas represent a standard deviation.}
\label{fig:exp_result}
\end{figure*}

\section{Sim-to-real Transfer}
\balance
Sim-to-real transfer has been a long-standing problem in robotics, due to the gap between the simulation environment and the real robot. In this section, we introduce several methods we used to deal with the sim-to-real transfer.
\subsection{Behavior Cloning}
In simulation environments, the robot can obtain complete environment information without any noise. However, in the real-world robot, the data obtained by the sensor may be too noisy to be used, and the robot can only determine the control signals with less sensor input. Specifically, we found that the robot we used for real-world experiments can only provide very noisy body velocity. To address the issue, an intuitive solution is to estimate the body velocity with the data from other sensors. However, introducing such a velocity estimation module will increase the complexity of our controller. Instead of explicitly estimating the body velocity, we employ the imitation learning \cite{ross2011reduction} to learn a student policy that determines control signals without the body velocity input, by cloning the behaviors of the teacher policy that has the velocity information. We test the new policy in the simulation environment, and it can achieve similar performance as the original policy.

\subsection{Domain Adaptation}
To overcome the sim-to-real gap, previous work also utilizes domain randomization to train a robust policy to adapt to a wide range of dynamic parameter settings \cite{peng2018sim}. However, recent work argues that it is possible to transfer the simulation controller directly with domain adaptation, by calibrating the dynamic parameters in simulation \cite{xie2020dynamics}. In this work, we use domain adaptation to narrow the sim-to-real gap instead. We control the robot with predefined control signals and collect the joint angles of each leg. Then we use black-box optimization to search the parameters in simulation that can minimize the average difference of the joint angles between the simulation and real-world robots. The parameter space we searched in the simulation environment contains the following factors:
\begin{itemize}
    \item Control latency $t_c$
    \item Foot friction $f_{foot}$
    \item {Base mass $m_{base}$}, Base inertia $I_{base}$
    \item Leg mass $m_{leg}$, Leg inertia $I_{leg}$
    \item Motor Kp $k_{p}$, Motor Kd $k_{d}$ for the PD controller
\end{itemize}
\subsection{Observation Randomization}
The observation obtained by the sensors is quite noisy in real-world deployment, which can degrade the controller performance in real robots. To mitigate this problem, we follow the prior work \cite{peng2020learning} and add noise into the sensor input while learning a student policy in the behavior cloning process. The noisy observation encourages the student policy to adopt robust behaviors in real-world deployment.


\begin{table*}[]
\centering
\caption{Experimental result}
\label{tab:my-table}
\resizebox{\textwidth}{!}{%
\begin{tabular}{cccccccccc}
\hline
  & SlopeSlope            & SlopeStair            & StairStair            & StairSlope            & Stair13              & Terrain          & Balance               & Gallop                & Cave                   \\ \hline
CPG-RBF & 116.2 \textpm 12.0            & 124.7 \textpm 10.0            & 113.9 \textpm 3.3             & 116.8 \textpm 3.1             & 88.5 \textpm 19.5            & 163.0 \textpm 3.4            & 4.5 \textpm 6.1               & 2266.5 \textpm 102.4          & -33.2 \textpm 16.2             \\
SAC     & 89.9 \textpm 4.8              & 79.1 \textpm 9.3              & 73.1 \textpm 11.9             & 80.9 \textpm 5.8              & 65.0 \textpm 3.3             & \textbf{2024.3 \textpm 173.5}         & 11.2 \textpm 2.8              & 1776.8 \textpm 1179.9         & -72.4 \textpm 36.6             \\
ETG-RL  & \textbf{2100.6 \textpm 570.4} & \textbf{1241.0 \textpm 731.4} & \textbf{1462.0 \textpm 355.5} & \textbf{1542.4 \textpm 156.7} & \textbf{763.5 \textpm 376.0} & 2003.9 \textpm 41.7          & \textbf{1901.3 \textpm 1097.7}         & \textbf{4465.5 \textpm 514.2} & \textbf{4130.9 \textpm 1825.1} \\ \hline
\end{tabular}%
}
\end{table*}

\section{EXPERIMENTS}
To evaluate the ETG-RL algorithm , we first compare it with the prior learning-based approaches in 9 challenging simulation tasks developed on the pybullet environment \cite{coumans2016pybullet}, a physics simulator for robotics. The baseline methods include central pattern generator with radial basis function (CPG-RBF) \cite{thor2020generic} and soft actor-critic (SAC) \cite{haarnoja2018soft}. We then present the ablation study to understand the contribution of each improvement proposed in ETG-RL. Finally, we test the performance of the learned controller in real-world tasks and deploy it on a 12-DoF quadrupedal robot.

\begin{figure}[b] 
\centering
\subfloat[]{\includegraphics[width=0.5\linewidth]{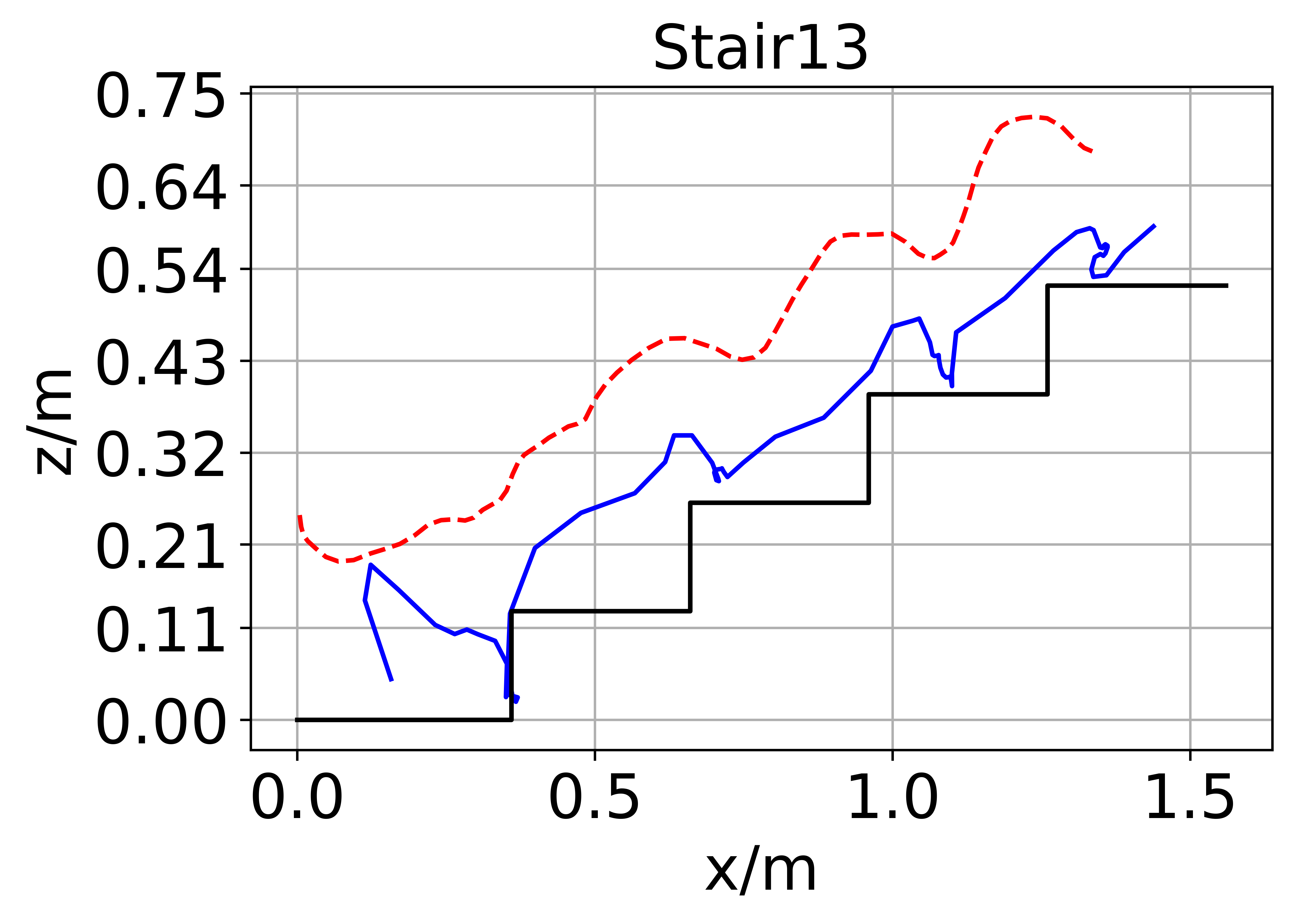}}
\subfloat[]{\includegraphics[width=0.5\linewidth]{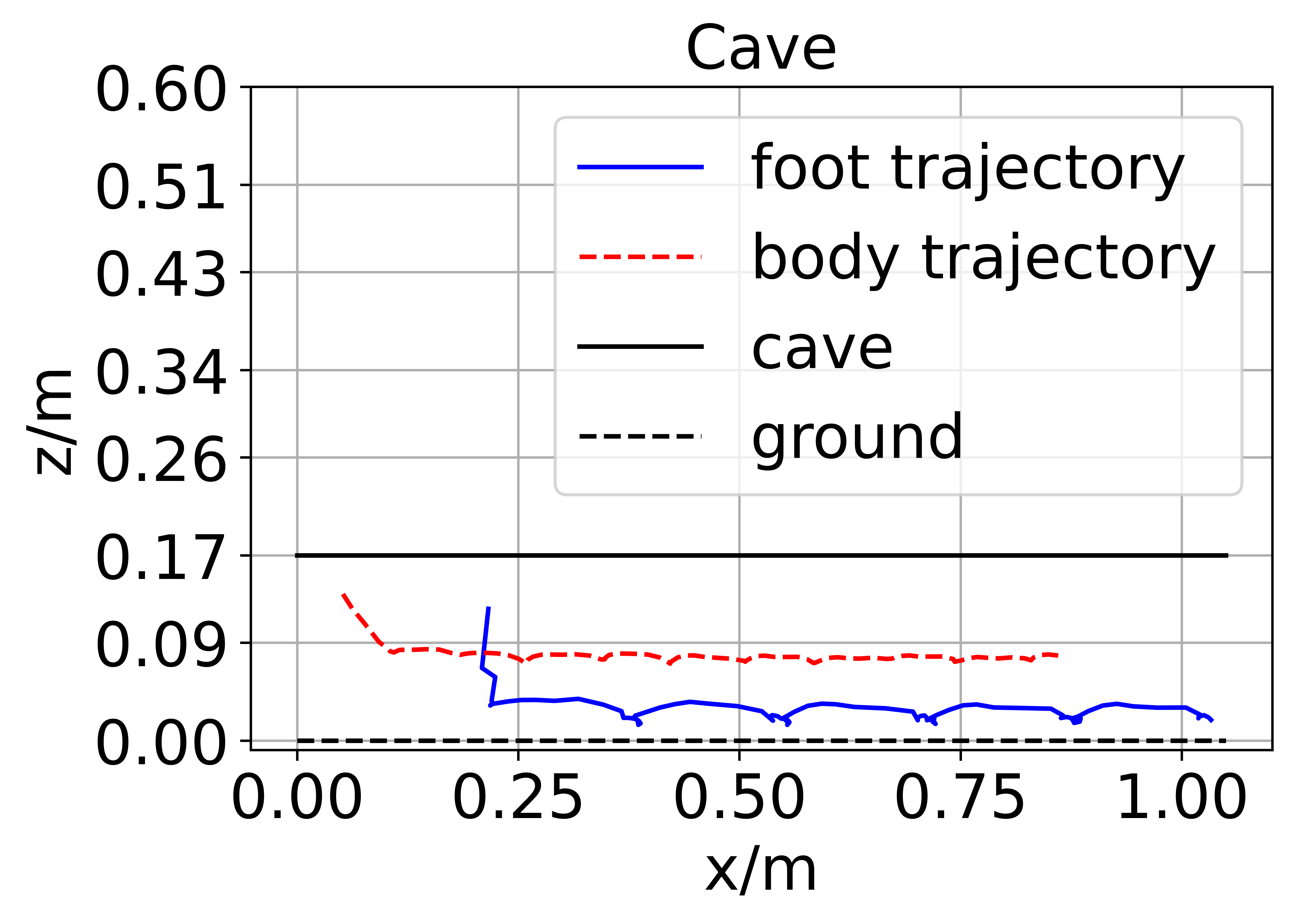}}
\caption{Body and foot trajectories of the robot in \textit{Stair13} and \textit{cave}. (a) \textit{Stair13} environment (b) \textit{cave} environment.}
\label{fig:foot_trajectory}
\end{figure}

\subsection{Simulation Experiments}
We developed 9 simulation tasks based on pybullet \cite{coumans2016pybullet} for evaluation, as shown in Figure~\ref{fig:env}. The tasks consist of challenging scenarios that require different locomotion skills, including climbing up and down stairs and slopes, traversing the uneven terrain, walking on a narrow balance beam, jumping over planks with wide gaps, and crawling through the cave.

The training curves are shown in Figure~\ref{fig:exp_result} and the result is presented in Table~\ref{tab:my-table}. The ETG-RL algorithm is the only algorithm that can solve all the tasks, and it significantly outperforms the baseline approaches in all tasks except the Terrain environment, where the robot is not prone to fall. Though prior work has shown that CPG-RBF based controller can walk stably on the flat land, it fails to traverse the uneven environments, and we found that it is difficult for CPG-RBF to recover balance while tripping over. SAC can solve the \textit{Terrain} task, as it is less possible to fall over, compared with other environments such as \textit{Balance}. The experimental result shows that as we employ the ETG to provide effective motion prior to guide the policy learning, the neural network policy can learn efficient gaits via RL. 
We also plot the foot trajectories of our controller in \textit{Stair13} and \textit{Cave} in Figure~\ref{fig:foot_trajectory}. The figures show that learning from scratch, the robot learns to raise its legs to climb up the stairs, and lower the body to climb through the cave, while the baseline approaches fail to solve these tasks.



\subsection{Ablation Study}
We design the ablation experiments to investigate the following questions: (1) Do ETG perform better than CPG in optimizing foot trajectory?  (2) Are evolutionary trajectory generator (ETG) and dual training essential for learning efficient gaits?
We define two variants of ETG-RL:
\begin{itemize}
    \item \textbf{CPG-RL} that replaces ETG with CPG, searching the optimal trajectory in parameter space rather than trajectory space.
    \item \textbf{TG-RL} that replaces the evolutionary trajectory generator with a fixed trajectory generator.
\end{itemize}
Note that in both CTG-RL and TG-RL, we initialize all the trajectory generators with the same parameters to ensure that they have the same motion prior.

We evaluate the variants in 9 tasks, and the experimental result is presented in Table~\ref{tab:ablation}. The performance degradation of CPG-RL shows that ETG can find better foot trajectories than CPG. As we search the foot trajectory in trajectory space, it is less likely to sample unexpected trajectories that dramatically deviate from the current trajectory, compared with sampling in parameter space. The second question can be answered by referencing the comparison between TG-RL and ETG-RL. The performance of ETG-RL drops significantly while we remove the dual training in almost all tasks except \textit{Terrain} and \textit{Balance}. For these two tasks, the motion prior provided by the trajectory generator has exhibited decent performance. The full algorithm is the only algorithm that solves all tasks.

\begin{table}[]
\centering
\caption{Ablation experiment}
\label{tab:ablation}
\resizebox{0.5\textwidth}{!}{%
\begin{tabular}{@{}cccc@{}}
\toprule
             &CPG-RL           & TG-RL       & ETG-RL                \\ \midrule
SlopeSlope  & 27.8\textpm18.1    & 1024.4\textpm248.8      & \textbf{2100.6\textpm570.4}  \\
SlopeStair  & 67.4\textpm30.6    & 411.5\textpm84.7       & \textbf{1241.0\textpm731.4}  \\
StairStair  & 71.0\textpm105.1   & 848.6\textpm225.8    & \textbf{1462.0\textpm355.5}  \\
StairSlope  & 58.3\textpm58.9    & 1203.8\textpm138.1    & \textbf{1542.4\textpm156.7}  \\
Stair13     & 6.2\textpm17.3     & 91.2\textpm6.9     & \textbf{763.5\textpm376.0}   \\
Terrain & 2046.5\textpm58.4  & \textbf{2060.1\textpm26.4}   & 2003.9\textpm41.7   \\
Balance     & 4.5\textpm7.2      & \textbf{2134.7\textpm627.5}     & 1901.3\textpm1097.7 \\
Gallop      & 1525.3\textpm291.8 & 3196.3\textpm563.8  & \textbf{4465.5\textpm514.2}  \\
Cave        & -32.2\textpm14.0   & 3567.5\textpm1338.8  & \textbf{4130.9\textpm1825.1} \\ \bottomrule
\end{tabular}%
}
\end{table}

\subsection{Physical Experiments}
As mentioned in the sim-to-real transfer, we train a student policy that does not directly rely on the noisy estimated velocity, by imitating the teacher policy that has the full sensor input in simulation. We then set up 6 real-world environments (Figure~\ref{fig:env}) that are similar to the simulation task to evaluate the controller performance in real world. We deploy the controller in a 12-DoF quadrupedal robot made by Unitree Robotics with the sim-to-real transfer, and the robot can successfully traverse all the environments, climbing up and down stairs and slopes, walking over the balance beam. To the best of our knowledge, this is the first learning-based controller that can walk over the balance beam in the real world. We also provide a video recording the controller performance in real-world tasks.

\section{CONCLUSIONS}
In this work, we present the ETG-RL algorithm for learning quadrupedal locomotion skills. We first devise an evolutionary trajectory generator to guide policy learning. The trajectory generator optimizes its foot trajectory efficiently by directly searching in the foot trajectory space, and thus provides effective prior on motion generation. To stabilize the training, we alternatively train the neural network policy via RL and optimize the trajectory generator. The experimental result in simulation shows that our approach can learn efficient gaits in a range of challenging environments such as walking over a narrow balance beam. We further validate the controller on a 12-DoF quadrupedal robot, and it can solve the real-world task with the sim-to-real transfer. We hope our work can facilitate the real-world applications of reinforcement learning in quadrupedal robots.


\addtolength{\textheight}{-12cm}   







{\small
\bibliographystyle{plain}
\bibliography{myref}
}

\end{document}